%% file: main.tex
\documentclass[10pt,letterpaper]{article}

\usepackage{cogsci}

\cogscifinalcopy 

\usepackage{pslatex}
\usepackage{apacite}
\usepackage{float} 

\usepackage{graphicx} 
\usepackage{xcolor}
\usepackage{amsmath, amsfonts, amssymb}
\usepackage{algorithm}
\usepackage[noend]{algpseudocode}

\title{The Delusional Hedge Algorithm as a Model of \\ Human Learning from Diverse Opinions}

\author{
    \large \bf Yun-Shiuan Chuang \\
      University of Wisconsin-Madison\\
    \texttt{yunshiuan.chuang@wisc.edu} \\   
    \And
    \large \bf Xiaojin Zhu\\   
    University of Wisconsin-Madison\\
    \texttt{jerryzhu@cs.wisc.edu} \\   
    \And
    \large \bf Timothy T. Rogers\\ 
    University of Wisconsin-Madison\\
    \texttt{ttrogers@wisc.edu}
}

\definecolor{delusion}{HTML}{ef8843}
\definecolor{hedge}{HTML}{5d9dd5}
\definecolor{accuracy}{HTML}{9f73bf}
\definecolor{trust}{HTML}{81be5b}
\definecolor{near}{HTML}{000099}
\definecolor{middle}{HTML}{3333ff}
\definecolor{far}{HTML}{9999ff}

\begin{document}

\maketitle

\input{sections/00_abstract}
\input{sections/01_introduction}

\input{sections/02_preliminaries}

\input{sections/03a_exp1_methods}
\input{sections/03b_exp1_results}

\input{sections/04a_exp2_methods}
\input{sections/04b_exp2_results}

\input{sections/05_discussion}

\input{sections/acknowledgements}

\bibliographystyle{apacite}

\setlength{\bibleftmargin}{.125in}
\setlength{\bibindent}{-\bibleftmargin}

\bibliography{main}

\end{document}

%% file: sections/00_abstract.tex
\begin{abstract}

Whereas cognitive models of learning often assume direct experience with both the features of an event and with a true label or outcome, much of everyday learning arises from hearing the opinions of others, without direct access to either the experience or the ground truth outcome. We consider how people can learn which opinions to trust in such scenarios by extending the {\it hedge algorithm}: a classic solution for learning from diverse information sources. We first introduce a semi-supervised variant we call the \textit{delusional hedge} capable of learning from both supervised and unsupervised experiences. In two experiments, we examine the alignment between human judgments and predictions from the standard hedge, the delusional hedge, and a heuristic baseline model. Results indicate that humans effectively incorporate both labeled and unlabeled information in a manner consistent with the delusional hedge algorithm---suggesting that human learners not only gauge the accuracy of information sources but also their consistency with other reliable sources. The findings advance our understanding of human learning from diverse opinions, with implications for the development of algorithms that better capture how people learn to weigh conflicting information sources.


\end{abstract}

%% file: sections/01_introduction.tex
\section{Introduction}
\label{sec:introduction}

Cognitive approaches to knowledge acquisition often propose that learning entails accurate supervision: on at least some learning trials, people observe a stimulus $x$, generate a predicted outcome $\hat{y}$, then receive true, accurate information about the correct outcome $y$. Much of everyday experience, however, involves neither direct experience of event features ($x$) nor exposure to ground-truth labels ($y$). Instead, learning arises from exposure to diverse and sometimes contradictory opinions expressed by other individuals, any of whom may be mistaken or deceptive---consider, for instance, reading on social media about whether vaccines are safe, whether members of a political party are dishonest, or whether global warming is a hoax. In such cases opinions may diverge radically, and exposure to both event features and ground truth may be exceedingly sparse. How then do people decide which information sources they should trust when updating their own beliefs?

In machine learning, the {\it hedge algorithm} provides a useful starting point for addressing this question \shortcite{freund1999adaptive, freund1997decision,mourtada2019optimality,cesa2007improved, auer2002adaptive}. In the typical setup \footnote{Machine learning uses the terms "experts," "predictions," and "advice" to describe the learning problem, often termed ``online learning from expert advice''\shortcite{cesa2006prediction,bousquet2002tracking,littlestone1994weighted}. We are interested in cases where information sources may have little expertise or may even be duplicitous, so we use the term "source" instead of "expert," and "opinion" instead of "prediction" or "advice"---hence ``online learning from diverse opinions.''}, the model agent does not view the stimulus features $x$ of a given event, but instead receives opinions about the event label from each of $k$ information sources, all varying in their accuracy or reliability. The agent must infer the correct label from the opinions offered. During learning, the $k$ opinions are presented together with the ground-truth label, and this information is used to update the weights (``trust'') given to each source. The hedge algorithm provides a means of updating weights that is optimal in the sense that it guarantees low bounds on learner {\it regret}, i.e., the difference between the sequence of decisions the agent makes over the course of learning and the best possible sequence of decisions it could have made had the most accurate information source been known from the outset. 

When it comes to understanding human behavior, however, the hedge algorithm is limited, because it is fully supervised---it still requires exposure to the ground truth label $y$ on each learning episode (or more technically the true loss for the episode which usually depends on the ground truth label).
In many cases, such exposure is sparse or non-existent: opinions from others vastly outnumber immediate experiences of the ground truth. That is, human learning is semi-supervised, or even unsupervised at times \shortcite{zhu2007humans,kalish2011can,gibson2013human,latourrette2019little,broker2022unsupervised,chuang2021using}. The current paper develops a semi-supervised variant of the hedge learning algorithm in which (a) the loss on supervised trials is exactly as specified by the standard hedge but (b) on unsupervised trials, the learner generates a ``delusional loss'' based on source agreement and source weights, then uses that delusional loss to update weights across information sources. We refer to this algorithm as the {\em delusional hedge}, because the learner effectively hallucinates the non-existent loss on unsupervised trials. We then describe two experiments in which human behavior in a simple learning task is compared to predictions of the standard hedge model, the delusional hedge, and a third heuristic model. The results strongly suggest that human learning from diverse and contradictory opinions is semi-supervised, in ways well-captured by the delusional hedge model.

%% file: sections/02_preliminaries.tex
\section{Preliminaries}
\label{sec:preliminaries}

\subsection{Online SSL with Source Opinion}
\label{subsec:online_learning_setting}

We consider an online semi-supervised learning (SSL) framework where an agent must learn a binary classification from a set of $K$ information sources (typically called ``experts'' in machine learning) within a time horizon $T$. Each source has a fixed decision boundary unknown to the agent, denoted as $\theta_1, \ldots, \theta_K$. At each time step $t$, an instance $(x,y)$ is drawn from the environment $(x_t, y_t) \sim P_{XY}$. Each source $k$ then provides a binary {\it opinion} about the category label $b_{tk}$ based on its respective decision boundary $\theta_k$ (i.e., $ b_{tk} = -1 \text{ if } x_t < \theta_k \text{ else } 1$).  The agent, without access to the instance $x_t$, must rely only on these $k$ opinions to form a prediction $\hat{y_t}$ about the true label. After generating a prediction, the agent may or may not view the ground-truth label $y^*$. \textit{Label visibility} at each time step is denoted by $v_1, v_2, \ldots, v_T$, where $v_t \in \{0, 1\}$. When $v_t = 1$, the trial is labeled; otherwise, when $v_t = 0$, the trial is unlabeled. The learning settings can be characterized as follows: \textit{Fully supervised setting}: When $\sum_{t=1}^{T} v_t = T$, all trials are labeled. \textit{Fully unsupervised setting}: When $\sum_{t=1}^{T} v_t = 0$, all trials are unlabeled. \textit{Semi-supervised setting}: When $0 < \sum_{t=1}^{T} v_t < T$, some trials are labeled and some are not.

\subsection{Online Learning with Diverse Opinions Algorithms}
\input{algorithms/algo_delusional_hedge}

\textbf{Hedge Algorithm.}
The \textit{hedge algorithm} (Algorithm~\ref{algo:delusional_hedge}, omitting blue text) is a classic method in online learning wherein a learner iteratively updates its \textit{trust} in a set of sources \cite{freund1999adaptive, freund1997decision}. At each time step $t$, the learner assigns trust $p_{tk}$ to each source $k$ based on their cumulative losses (step 2). After the sources provide their opinions $(b_{t1},\dots,b_{tK})$ (step 3 and 4), the learner summarizes these (step 5) and makes its own prediction $\hat y_t$ (step 6) based on its trust $p_{tk}$ in each source. When a ground-truth label $y_t$ is revealed ($v_t=1$), the learner updates the cumulative loss for each source based on their predictions, using a 0-1 loss function $1[b_{tk}\neq y_t]$ (step 7). At the next time step, the learner updates its trust $p_{tk}$ in each source based on the cumulative loss using a softmax function with the learning rate $\eta$ (step 2). In \textit{standard hedge}, the learner does not update its trust in the different sources when the label $y_t$ is not present ($v_t=0$). Thus the agent learns whom to trust from supervised trials only, adjusting its behavior accordingly from trial to trial. Theoretical analysis shows that standard hedge has a bound on the worst-case regret of order $O(\sqrt{T logK})$ if the learning rate $\eta$ is properly chosen \cite{mourtada2019optimality,cesa2007improved, auer2002adaptive}.

\textbf{Delusional Hedge Algorithm.}
To accommodate unlabeled trials, we propose the \textit{delusional hedge algorithm} shown in Algorithm~\ref{algo:delusional_hedge} with blue text included. The key difference lies in step 7: when the true label $y_t$ is not revealed ($v_t=0$) the delusional hedge computes a \textit{delusional loss} (${q_{t, -b_{tk}}}$) for each source, defined as the sum of trust across all sources with {\it opposite} predictions (i.e., for a source with prediction $b_{tk}$, the delusional loss $q_{t,-b_{tk}} = \sum_{k': b_{tk'}=-b_{tk}} p_{tk'}$). This is equivalent to updating the trust based on the expected 0-1 loss if the true label were drawn from a Bernoulli distribution $Ber(q_{t,1})$. The delusional loss is weighted by a free hyperparameter $\alpha>0$ so the agent can weight information from labeled and unlabeled instances differently. When $\alpha=0$, this reduces to the standard hedge algorithm. Intuitively, the idea is that, if the sources that disagree with a given source $k$ collectively are highly trusted, then the weight on $k$ should change a lot; if the disagreeing sources are {\it not} collectively highly trusted, it should not change very much. Note that a large change can accrue when (a) trust is relatively uniform but many sources disagree with $k$, or (b) few sources disagree with $k$ but at least one is very highly trusted. Thus, the delusional hedge provides a way of exploiting both consensus amongst sources and high trust in specific sources when learning from unlabeled trials.


\textbf{Accuracy-Majority Heuristic.}
In the following experiments we compare the two hedge variants to one another and to a heuristic \textit{accuracy-majority} model where (1) the agent follows the source with the highest cumulative accuracy at each time step; (2) when sources share the same top accuracy, the agent follows the source whose prediction is most frequently in the majority (i.e., has the highest ratio of being in the majority across all trials); and (3) in rare cases where ties still persist after considering both accuracy and majority counts, the subject randomly selects one source to follow. While this heuristic baseline lacks free parameters and thus cannot be compared with hedge variants via likelihoods, we will see that it generates distinct predictions that provide a useful contrast to standard and delusional hedge algorithms.

%% file: algorithms/algo_delusional_hedge.tex
\begin{algorithm}[H]
	\begin{algorithmic}[1]
        \small
		\caption{\textcolor{blue}{(Delusional)} Hedge Algorithm}
        \label{algo:delusional_hedge}
		\Require horizon $T$, learning rate $\eta$, $K$ sources $\theta_1, \ldots, \theta_K$, $P_{xy}$, label visibility $v_1, v_2, \ldots, v_T$, weight of delusional loss $\alpha$
		\For{\(t = 1, 2 \dots ,T\)}
			\State {Learner assigns trust to sources} 
   
				$p_{tk} = 
				{
				\exp\left(-\eta \sum_{\tau=1}^{t-1}\ell_{\tau k} \right)
				\over
				\sum_{i=1}^K \exp\left(-\eta \sum_{\tau=1}^{t-1}\ell_{\tau i} \right)
				}, \; \forall k\in[K]$
				
    \Comment When $t=1$, $p_{tk}=1/K$ because the sum is empty.
			\State {Environment draws an item and its label $(x_t, y_t) \sim P_{XY}$}
			\State {Sources predict their labels} 
				$\forall k, \;\;\; b_{tk} = \left\{
				\begin{array}{ll}
				-1, & x_t < \theta_k \\
				1, & x_t \ge \theta_k
				\end{array}
				\right.$
			\State {Learner summarizes source predictions}
   
				$q_{t,-1} := \sum_{k: b_{tk}=-1} p_{tk}, \;\;\; q_{t,1} := \sum_{k: b_{tk}=1} p_{tk}$
			\State {Learner makes label prediction $\hat y_t \sim Ber(q_{t,1})$}
			\State {If ground-truth label $y_t$ is available, updates with 0-1 loss; \textcolor{blue}{otherwise, the learner hallucinates loss:}}
				
            $\ell_{tk} = \left\{ \begin{array}{ll} 1[b_{tk}\neq y_t]  & \mbox{label $y_t$ given ($v_t=1$)} \\ \textcolor{blue}{\alpha \times q_{t, -b_{tk}}} & \textcolor{blue}{\mbox{label $y_t$ not given ($v_t=0$)}} \end{array} \right., \forall k$
            
            \Comment This is the expected 0-1 loss (weighted by $\alpha$) to source $k$, as if the true label were drawn from $Ber(q_{t,1})$.
		\EndFor 
	\end{algorithmic}
\end{algorithm}

%% file: sections/03a_exp1_methods.tex
\section{Experiment 1}
\label{sec:exp1_methods}
\input{figures/human_exp_ui_merged}

\input{figures/exp1_follow_pos1_each_trial_type_over_time_by_group_vs_agents}



\textbf{Behavioral Experiment Procedure.} To evaluate the different models we devised an experimental procedure capturing key elements of the learning scenario. Participants imagined they were stranded on a deserted island together with three other survivors (``sources''), each providing advice on how best to consume the native fruits, which could either be eaten fresh or turned into jam. The fruits varied in shape along a one-dimensional manifold $x \in [0, 300]$, and the corresponding category (fresh or in jam, $y \in \{-1,+1\}$) was determined by their position on the manifold relative to a true decision boundary ($\theta^* = 150$) unknown to the participant (Figure~\ref{fig:human_exp_ui_merged}a). The joint probability distribution $ (x,y) \sim P_{xy} $ determined the quality of the fruit. In this experiment, $x$ followed a uniform distribution, $ x \sim P_x = U(0,300) $, while the outcome $ y $ is determined based on whether $ x $ is less than or greater than $ \theta^* $. If $ x < \theta^* $, the correct decision is to eat the fruit fresh ($ y = -1 $), otherwise, the fruit should be turned into jam ($ y = +1 $). 

All fruits were hidden in identical boxes and invisible to the participant (Figure~\ref{fig:human_exp_ui_merged}b); instead, the three sources ``looked in the box'' and provided an opinion about the category based on their own unique category boundary, with one source using a boundary far from the ground truth  ($\theta_{far} = 50$), one near the ground truth ($\theta_{near} = 165$), and one halfway between these ($\theta_{middle} = 107.5$). The Near source, possessing a boundary closest to $\theta^*$, provided the most accurate predictions, followed by the Middle source. The Middle source's advice always agreed with at least one other source, so this source was most frequently in the majority. Additionally, the experiment counterbalanced the face images of the sources, their positional order, and the associations between $-1/+1$ and the verbal labels of "jam" or "fresh" across participants.

Each trial began with a {\it prediction phase} (Figure~\ref{fig:human_exp_ui_merged}b) in which participants made a decision $\hat{y_t}$ based on the three sources' opinions ($b_{tk}$), followed by a {\it feedback phase} in which, if the trial was labeled (Figure~\ref{fig:human_exp_ui_merged}c), the label $y_t$ was revealed by "source chef." In the unlabeled trials (Figure~\ref{fig:human_exp_ui_merged}d), no label $y_t$ was presented. After 100 such trials, we measured participants' trust in each source, first asking participants to select the most accurate source, then to choose the source most often in the majority, and finally to rate each source's 1) knowledgeability, 2) accuracy, 3) trustworthiness, and 4) attractiveness on a slider-based scale from $-100$ (``Not at All'') to $100$ (``Absolutely''). Attractiveness was included as a control rating task that should not be affected by trust learning.

\textbf{Conditions of Different Supervision Levels.}
Label visibility of each trial, denoted as $ v_1, v_2, \ldots, v_T $, was stochastically determined by the parameter $p_{visible}$ which specifies the probability that a trial's label will be visible to the participants, i.e., $ v_t \sim Ber(p_{visible})$. We created five between-subject conditions with different levels of supervision, ranging from fully unsupervised ($ p_{visible} = 0$) to fully supervised ($ p_{visible} = 1 $). Intermediate values of 1/4, 2/4, and 3/4 correspond to semi-supervised conditions. 

\textbf{Participants.}
Participants were undergraduates at a university who participated in exchange for course credit.  The study was approved by the Institutional Review Board (IRB). 186 students were recruited, with 181 completing the experiment. The participants were randomly assigned to one of the five supervision conditions: 33 in condition $p_{visible} = 0$, 38 in condition $p_{visible} = 0.25$, 35 in condition $p_{visible} = 0.5$, 38 in $p_{visible} = 0.75$, and 37 in condition $p_{visible} = 1$. 

\textbf{Model Fitting.}
To fit the learning algorithms to human data, we tuned the hyperparameters $\eta$ (and $\alpha$, if applicable) for each participant using maximum likelihood estimation based on the participants' actual predictions $\hat{y_t}$ and the probabilities predicted by the algorithm $q_{t,1}$ (step 6 in Algorithm~\ref{algo:delusional_hedge}). 

%% file: figures/human_exp_ui_merged.tex
\begin{figure}[tbh!]
    \centering
    \includegraphics[width=1\linewidth]{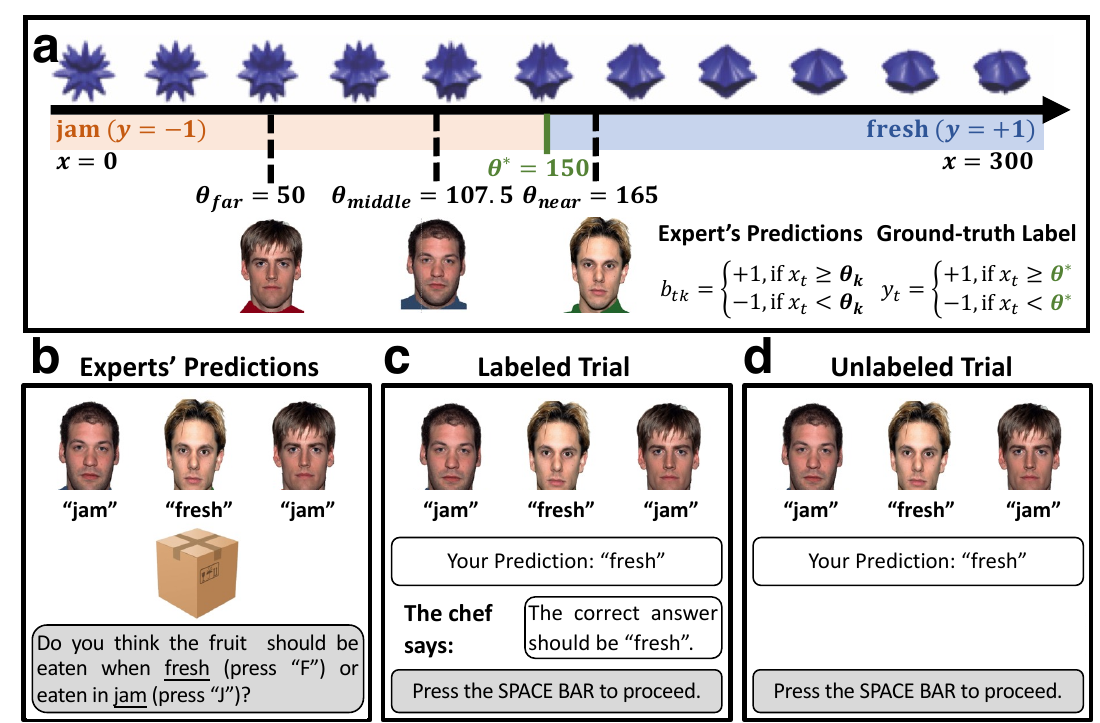}
    \caption{Human experimental setup. Panel (a) shows a 1D space $[0,300]$ where ``fruits'' are placed, with three sources having different decision boundaries $\theta$, and the true boundary at $\theta^* = 150$. Panels (b)-(d) display the user interface: (b) shows the three sources providing their opinions on the hidden fruit, while the participant makes a prediction; (c) reveals the true label post-prediction in labeled trials; and (d) shows the display omitting the label feedback in unlabeled trials.}
    \label{fig:human_exp_ui_merged}
\end{figure}

%% file: figures/exp1_follow_pos1_each_trial_type_over_time_by_group_vs_agents.tex
\begin{figure*}[htb!]
    \centering
    \includegraphics[width=1\linewidth]{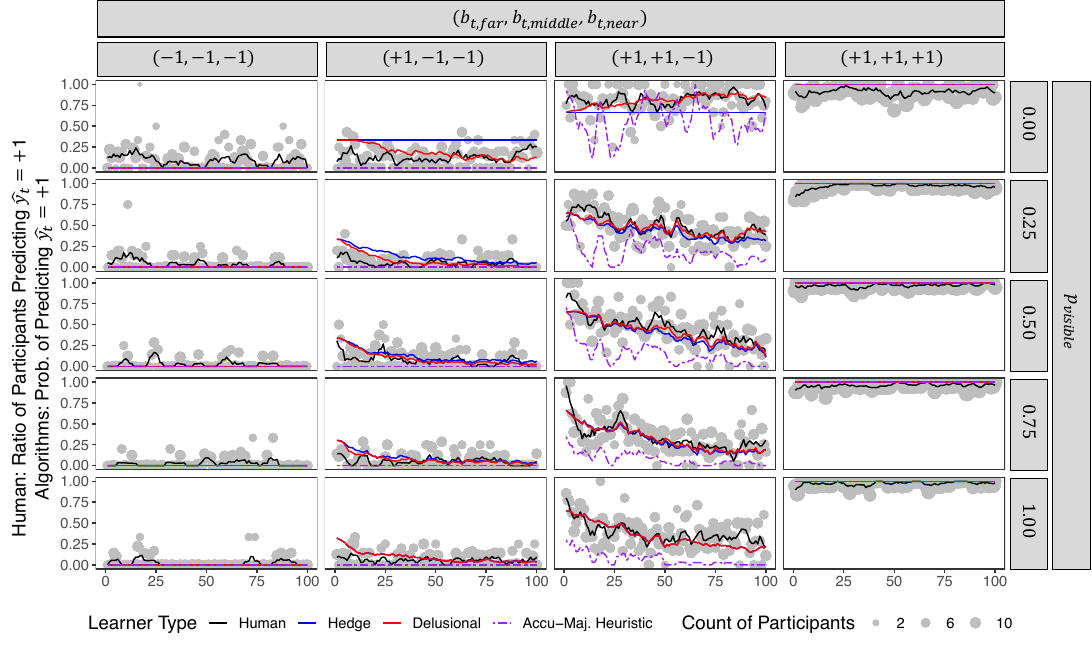}    
    \caption{Comparison of participant responses with the \textcolor{hedge}{standard hedge} algorithm, the \textcolor{delusion}{delusional hedge} algorithm, and the \textcolor{accuracy}{accuracy-majority heuristic} over 100 time steps. Each row represents a different level of supervision (from $p_{visible} = 0$ to $p_{visible} = 1$), and each column corresponds to one of the four unique combinations of source opinions $(b_{t, far}, b_{t, middle}, b_{t, near})$. Within each line plot, the black line shows the moving average of the ratio of participants predicting $\hat{y_t} = 1$ (y-axis) over the 100 time steps (x-axis), and the \textcolor{hedge}{blue}, \textcolor{delusion}{red}, and \textcolor{accuracy}{purple} lines represent the prediction probability of the standard Hedge algorithm, the delusional hedge algorithm, and the accuracy-majority heuristic, respectively, averaged across participants at each time step. Grey dots depict the proportion of participants with $\hat{y_t} = 1$ at each time step, with dot size denoting participant count.}

\label{fig:exp1_follow_pos1_each_trial_type_over_time_by_group_vs_agents}
\end{figure*}

%% file: sections/03b_exp1_results.tex
\subsection{Results of Experiment 1}
\label{sec:exp1_results}

\input{figures/exp1_rating_by_group}

\textbf{Online Learning Behavior.}
Figure~\ref{fig:exp1_follow_pos1_each_trial_type_over_time_by_group_vs_agents} shows human behavioral responses against different learning algorithms over time.  In the fully unsupervised setting (the first row of the figure), participants predominantly followed the majority opinion, predicting -1 when at least two sources predicted -1, and predicting +1 otherwise. This tendency was more accurately captured by the delusional hedge algorithm compared to the standard hedge algorithm--perhaps trivially, since the standard hedge does not learn at all in this condition. 
In the fully supervised condition (last row of the figure), when the majority of sources (Middle and Far) disagreed with the most accurate source (Near), i.e., $(b_{t,far},b_{t,middle},b_{t,far})=(1,1,-1)$ (third column), participants initially followed the majority but gradually shifted to align with the Near source. The standard and delusional hedge algorithms are equivalent in this fully supervised context and both mirrored this learning behavior. In the semi-supervised conditions ($0 < p_{visible} < 1$), participants exhibited a learning pattern similar to that in the fully supervised setting when faced with the same scenario in the third column, albeit with a flatter slope as $p_{visible}$ decreased. Both the delusional hedge and standard hedge algorithms showed qualitatively similar learning curves in these conditions. To test the relative fitness of each algorithm to human data, we conducted a likelihood ratio test, treating the standard hedge as a nested model within the delusional hedge algorithm (where $\alpha = 0$). Across all conditions, the results significantly favored the delusional hedge algorithm, $\lambda (144) = 740.5$, $p < .001$. Furthermore, likelihood ratio tests conducted for each semi-supervised condition consistently showed a better fit to the human data for the delusional than the standard hedge algorithm, $ps < .05$ (Bonferroni corrected). 
Finally, the accuracy-majority heuristic mode did not capture participant behavior well, showing a much stronger tendency to follow the Near source when $p_{visible}>0$, especially early in learning.


\textbf{Final-State Trust in Algorithm Simulations.}
The first row of Figure~\ref{fig:exp1_rating_by_group_human_vs_hedge} shows the final-state trust levels ($p_{Tk}$) assigned by both standard and delusional hedge algorithms. Mixed-effect ANOVAs revealed significant $p_{visible}$ $\times$ source interactions for both algorithms (Delusional hedge: $F(2,537)=123.28, p<.001$; Standard hedge: $F(2,537)=90.07, p<.001$). Post-hoc analysis showed distinct patterns: for delusional hedge, the Near source consistently received the highest trust for $p_{visible} > 0$, while the Middle source had the highest trust in the fully unsupervised setting ($p_{visible} = 0$) due to delusional loss ($p_{T,middle}=0.764$). Conversely, standard hedge algorithm assigned equal trust to all sources ($p_{Tk}=1/3$) in the unsupervised condition. Thus, the fully unsupervised condition highlighted a clear distinction between the standard hedge algorithm and the delusional hedge algorithm.

\textbf{Source Ratings.}
The participants' self-report ratings (second and third rows of Figure~\ref{fig:exp1_rating_by_group_human_vs_hedge}) were analyzed using a mixed-effect ANOVA (5 $p_{visible}$ conditions x 3 sources) for each rating type. When $p_{visible} > 0$, participants rated the Near source as most accurate, trustworthy, and knowledgeable, followed by the Middle source, $ps < .001$. Critically, in the fully unsupervised setting, the Middle source was rated highest for accuracy, trustworthiness, and knowledgeability, echoing the delusional hedge algorithm's behavior, $ps < .05$. Attractiveness rating showed no significant difference across sources, $F(8,352)=0.818, p=.587$.



The fourth row of Figure~\ref{fig:exp1_rating_by_group_human_vs_hedge} shows participant choices about which source was most accurate and which was most often in the majority. For accuracy, a mixed-effect (5 $p_{visible}$ conditions x 3 sources) ANOVA revealed (a) a significant main effect of source for each $p_{visible}$ condition ($ps<.001$) and (b) a significant $p_{visible}$ $\times$ source interaction ($\chi^2(8)=66.04, p<.001$): the Near source was chosen as most accurate when $p_{visible} > 0$, $ps<.001$, but the Middle source was chosen as most accurate when $p_{visible}=0$, $p<.001$. The majority choice, however, also showed a significant $p_{visible}$ $\times$ source interaction ($\chi^2(8)=20.92, p<.007$) with post-hoc analysis revealing that the Middle source was chosen significantly more often than the Near source only when $p_{visible} = 0, z = 3.94, p < .001$.



\textbf{Discussion.}
The delusional hedge algorithm better explained human data than the standard hedge algorithm, especially in the fully unsupervised setting. The accuracy-majority heuristic model did not explain human behaviors well. Although the delusional hedge also showed better fit to human data in the semi-supervised conditions (as suggested by the likelihood ratio tests), both algorithms yielded \textit{qualitatively similar} predictions in learning to increasingly trust the Near source---a similarity also observed in participants' ratings of / decisions about the different sources, where both algorithms assigned most trust to the Near source followed by the Middle source. Experiment 2 creates scenarios where the two algorithms make \textit{qualitatively distinct} predictions, allowing us to evaluate which algorithm aligns more closely with human behavior in semi-supervised settings. 



%% file: figures/exp1_rating_by_group.tex
\begin{figure}[htb!]
    \centering
    \includegraphics[width=1\linewidth]{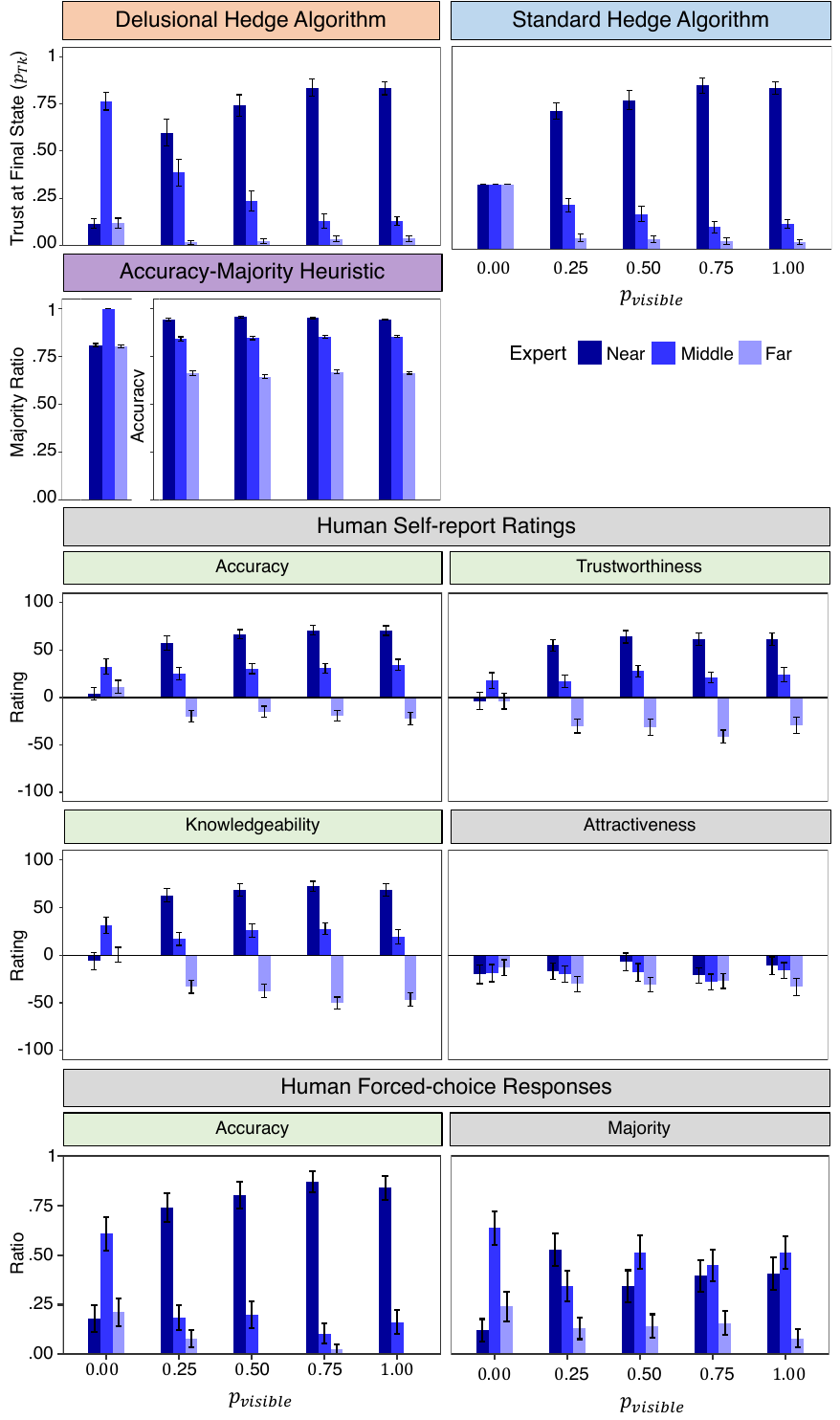}  
    \caption{Under five supervision levels ($0 \leq p_{visible} \leq 1$), the trust assigned to each source at the final state ($p_{Tk}$) by the \textcolor{hedge}{standard hedge} algorithm and the \textcolor{delusion}{delusional hedge} algorithm (top row), along with the accuracy (or majority ratio if accuracy is undefined) used by the \textcolor{accuracy}{accuracy-majority heuristic} (second row), as well as participants’ source ratings (third and forth rows), and proportion of times each source was chosen as most accurate or most often in the majority (bottom row). The bars for sources color-coded as \textcolor{near}{dark blue} (Near), \textcolor{middle}{blue} (Middle), and \textcolor{far}{violet} (Far). The ratings and responses designed to gauge participants' trust in sources are color-coded as \textcolor{trust}{green}. The error bars display the standard errors. }
\label{fig:exp1_rating_by_group_human_vs_hedge}
\end{figure}

%% file: sections/04a_exp2_methods.tex
\section{Experiment 2}


\label{sec:exp2_methods}


\textbf{Behavioral Experiment Procedure.}
The experiment comprised two phases. In the first, all participants saw the same five labeled trials in which the Near source always produced the correct label while and the other two always produced the incorrect label. The following 95 trials were unlabeled, split into two between-subject conditions. In the \textit{``M=F''} condition, the Middle source always produced the same opinion as the Far source, while in the \textit{M=N} condition, the Middle source always produced the same opinion as the Near source. The experiment then concluded with the same procedure for evaluating trust in the three sources from Experiment 1.

The rationale for the design is as follows: from the standard hedge algorithm, the five supervised trials should establish greater initial trust in the Near source, and the {\em same} lower amount of trust for the Middle and Far sources. Because the remaining trials are unsupervised, standard hedge should then predict that this is how trust will be allotted at the end of learning in both conditions: greater trust for the Near source and a lower and equivalent amount of trust for Middle and Far sources. The heuristic model should make the same prediction, because the Near source has a perfect track record of accuracy in the initial supervised trials (hence the highest cumulative accuracy). The delusional hedge, however, should show different patterns in the two conditions. When the Middle source always agrees with the Near source, the delusional loss should cause it to increase in trust on unsupervised trials--the higher trust in the Near source will ``bleed over'' to influence trust a second source that always agrees with the highly trusted source. When the Middle source always agrees with the Far source, however, there should be no difference in trust between Middle and Far. Thus, the pattern of trust observed in human learning can adjudicate these models.

\paragraph{Participants.}
We followed the same recruitment process as Experiment 1. A total of 80 students were recruited, with 77 completing the experiment. The participants were randomly assigned to one of the two unlabeled conditions: 39 in the \textit{``M=F''} condition, and 38 in the \textit{``M=N''} condition.

\paragraph{Model Fitting.}
We used the same model fitting procedure as in Experiment 1 to tune the hyperparameters $\eta$ (and $\alpha$, if applicable) for each participant using MLE.

%% file: sections/04b_exp2_results.tex
\subsection{Results of Experiment 2}
\label{sec:exp2_results}

\paragraph{Final-State Trust in Algorithm Simulations.}

\input{figures/exp2_rating_by_group}

Following our hypotheses, we focused on the trust $p_{Tk}$ assigned to the Middle and Far sources by both the standard and delusional hedge algorithms under the two unlabeled conditions. For the delusional hedge algorithm, a mixed-effect ANOVA on $p_{Tk}$ (2 unlabeled conditions $\times$ \{Middle \& Far\} sources) revealed a significant source $\times$ condition interaction: in the \textit{``M=N''} condition, the trust in the Middle source ($p_{T,Middle}=0.128$) was significantly higher than in the Far source ($p_{T,Far}=0.067$), $F(1,37)=16.03$, $p<.001$. In the \textit{``M=F''} condition,  the Middle and Far sources had the exact same trust because they always had identical predictions $b_{tk}$ (both with $p_{Tk}=0.07$). In contrast, for the standard hedge algorithm, another mixed-effect ANOVA indicated no significant source $\times$ unlabeled condition interaction as the Middle and Far sources had the same levels of trust (\textit{``M=N''} condition: $p_{T,middle}=p_{T,far}=0.13$; \textit{``M=F''} condition: $p_{T,middle}=p_{T,far}=0.10$).


\textbf{Source ratings.}
As shown in Figure~\ref{fig:exp2_rating_by_group}, a mixed-effect ANOVA (2 unlabeled conditions $\times$ \{Middle \& Far\} sources) was performed for each rating type. For accuracy, trustworthiness, and knowledgeability ratings, the interaction was significant, $ps<.05$, with post-hoc analysis showing that the Middle source was consistently rated as more accurate, trustworthy, and knowledgeable than the Far source in the \textit{``M=N''} condition ($ps<.01$) but not the \textit{``M=F''} condition ($ps>.05$). Finally, the condition $\times$ source interaction was not significant for attractiveness ($F(1,75)=0.22$, $p=.640$). Overall, participants' trust measured by self-report ratings aligned with the predictions of the delusional hedge algorithm.


\paragraph{Discussion.}
Experiment 2 suggests that human learners make use of unsupervised data when assigning trust to different information sources. In contradiction to predictions from the standard hedge, unlabelled trials increased the trust given to an initially-untrusted source that often agrees with a more-trusted source. Thus, more trust accrued to the Middle source than the Far source only when the Middle source frequently agreed with the Near source on unlabelled trials. This suggests that, in semi-supervised settings, human trust is influenced, not only by the supervised accuracy of a source, but also by the consistency of its opinions with the reliable source---a behavior predicted by the delusional hedge algorithm.


%% file: figures/exp2_rating_by_group.tex
\begin{figure}[tb!]
    \centering
    \includegraphics[width=1\linewidth]{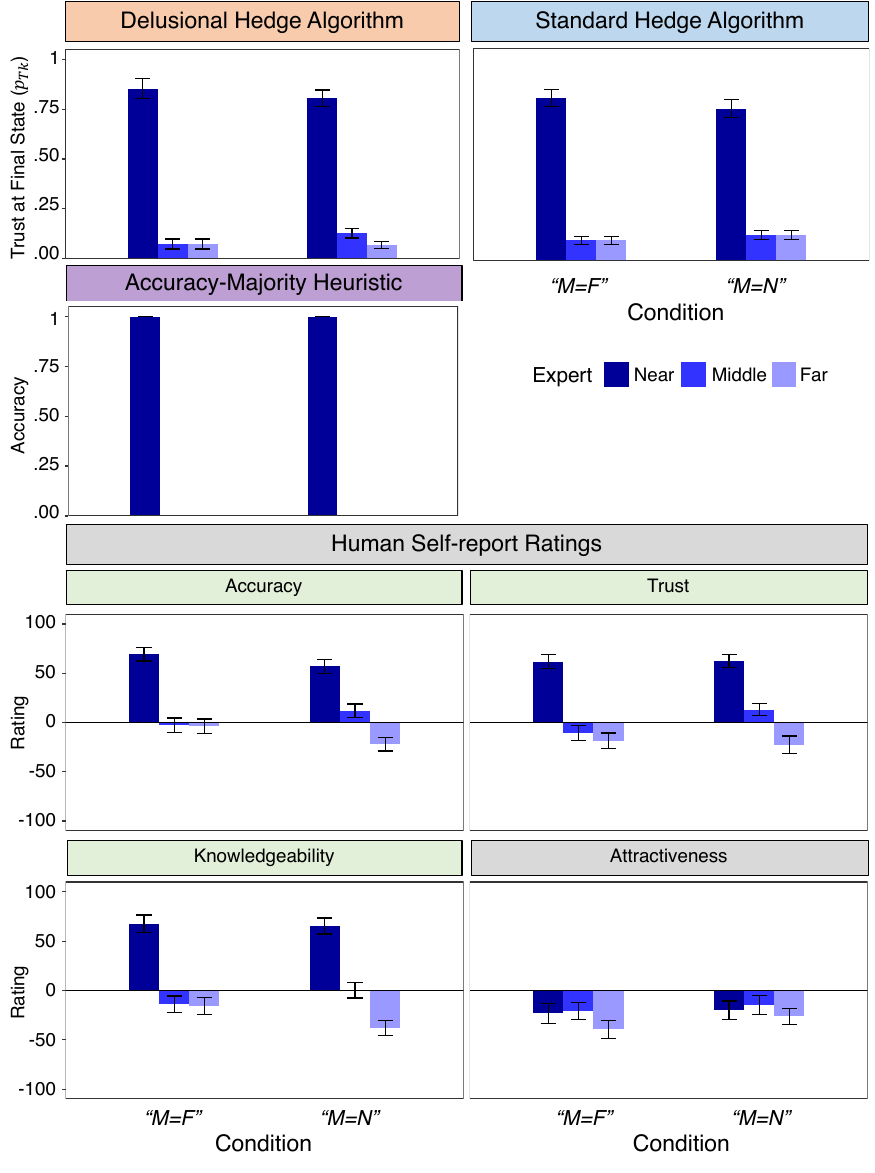}    
    \caption{Under the ``M=F'' and ``M=N'' conditions,  the trust assigned to each source at the final state ($p_{Tk}$) by the \textcolor{hedge}{standard hedge} algorithm and the \textcolor{delusion}{delusional hedge} algorithm (top row), along with the accuracy (or majority ratio if accuracy is undefined) used by the \textcolor{accuracy}{accuracy-majority heuristic} (second row), as well as participants’ source ratings (third and forth rows). The bars for sources color-coded as \textcolor{near}{dark blue} (Near), \textcolor{middle}{blue} (Middle), and \textcolor{far}{violet} (Far). The ratings designed to gauge participants' trust in sources are color-coded as \textcolor{trust}{green}. The error bars display the standard errors. }
    \label{fig:exp2_rating_by_group}
\end{figure}

%% file: sections/05_discussion.tex
\section{Conclusion}
\label{sec:discussion}

In many important real-world scenarios people learn, not from a directly observed-event and corresponding ground-truth label, but through experience with diverse and potentially contradictory opinions of others. Understanding how maladaptive beliefs emerge and persist in society requires computational formalisms that can characterize how people learn which opinions to trust in such scenarios. Machine learning provides a rich source of potential hypotheses in the form of models with well-understood properties and formal guarantees. We have focused on one such model, the hedge algorithm, showing how it can be adapted to semi-supervised situations that may better capture the reality of human learning. Our experiments showed that people integrate both labeled and unlabeled data when learning from diverse opinions, producing behaviors that align well with the predictions of the delusional hedge algorithm. The work provides an initial starting point for bridging computational learning theory and approaches to human social learning that, we hope, can be extended through formal analysis of the delusional hedge itself and through consideration of a broader range of approaches in both machine learning and cognitive science.


%% file: sections/acknowledgements.tex
\section*{Acknowledgements}

We thank the anonymous reviewers for their feedback. This project is supported in part by NSF grants 1545481, 1704117, 1836978, 2023239, 2041428, 2202457, ARO MURI W911NF2110317, and AF CoE FA9550-18-1-0166.